\documentclass{article}
\usepackage[margin=2cm]{geometry}
\usepackage[pdftex]{graphicx}
\usepackage{booktabs}
\usepackage{amsmath}
\usepackage{amssymb}
\usepackage{subfigure}
\usepackage{color}

\usepackage{paralist}
\usepackage{url}

\usepackage{amsmath}
\usepackage{amsfonts} 
\usepackage{algorithm2e}
\usepackage{tikz}
\usepackage{verbatim}
\begin{document}

\title{Intraoperative margin assessment of human breast tissue in optical coherence tomography images using deep neural networks}

\author{Amal~Rannen~Triki\thanks{A. Rannen Triki is with the Department of Computational Science and Engineering, Yonsei University, Korea and the Center for Processing Speech and Images, Department of Electrical Engineering, KU Leuven, Belgium - e-mail: amal.rannen@esat.kuleuven.be.},  Matthew~B.~Blaschko\thanks{M. B. Blaschko is with the Center for Processing Speech and Images, Department of Electrical Engineering, KU Leuven, Belgium - e-mail: matthew.blaschko@esat.kuleuven.be.}, Yoon~Mo~Jung\thanks{Y. M. Jung is with Sungkyunkwan University, Korea.}, Seungri~Song,\\
 Hyun~Ju~Han, Seung~Il~Kim, and~Chulmin~Joo\thanks{S. Song, H. J. Han, S. I. Kim, and C. Joo are with Yonsei University, Korea.}
}
\maketitle

\begin{center}
This work has been submitted to the IEEE for possible publication.
\end{center}

\begin{abstract}
\textit{Objective:} In this work, we perform margin assessment of human breast tissue from optical coherence tomography
(OCT) images using deep neural networks (DNNs). This work simulates an
intraoperative setting for breast cancer lumpectomy. 
\textit{Methods:} To train the DNNs, we use both the state-of-the-art methods
(Weight Decay and DropOut) and a newly introduced regularization method based on function norms. Commonly used
methods can fail when only a small database is available. The use
of a function norm introduces a direct control over the complexity of the function with the aim of diminishing the risk of overfitting.
\textit{Results:} As neither the code nor the data
of previous results are publicly available, the obtained results are
compared with reported results in the literature for a conservative
comparison. Moreover, our method is applied to locally collected
data on several data configurations. The reported results are the average over the
different trials.
\textit{Conclusion:} The experimental results show that
the use of DNNs yields significantly better results than other
techniques when evaluated in terms of sensitivity, specificity, F1
score, G-mean and Matthews correlation coefficient. Function norm regularization yielded higher and more robust results than competing methods.
\textit{Significance:} We have demonstrated
a system that shows high promise for (partially) automated
margin assessment of human breast tissue, Equal error rate (EER) is reduced from approximately 12\% (the lowest reported in the literature) to 5\%\,--\,a 58\%
reduction. The method is computationally feasible for intraoperative
application (less than 2 seconds per image).  
\end{abstract}

\textbf{Index:} Breast cancer, Margin assessment, Optical Coherence Tomography (OCT), Deep Neural Network (DNN), Weight Decay, DropOut, Function norm regularization

\section{Introduction}

Breast cancer represents one of the leading causes of cancer deaths among women. It has been estimated that 252,710 new cases and 40,610 deaths would occur in the United States in 2017~\cite{usCancerFigures}. Once diagnosed as breast cancer, 60\% to 70\% of patients undergo breast-conserving surgery or lumpectomy for surgical resection of the tumor~\cite{usCancerFigures2}. Alternative noninvasive strategies may also be employed such as MRI-guided focused ultrasound surgery~\cite{huber2001new}, but lumpectomy remains the preferred method of treatment. 

At present, resected breast tissues are examined postoperatively by pathologists to demarcate tumor margins. The procedure involves analyzing the gross resection specimen macroscopically and representative histopathology sections of the tissue through microscopy. Yet, it is not practical to evaluate the entire tumor margin with this method, as a large amount of histologic tissue sections need to be examined to fully evaluate a breast tissue specimen. Pathologists therefore analyze the regions of interest selected from macroscopic evaluation, which may lead to sampling error. In 20\% to 60\% of cases, additional tumor is found at the margins~\cite{gwin1993incidence}, and additional surgery is required, which results in higher treatment cost, greater morbidity, infection risk, and delayed adjuvant therapy~\cite{davis2013topical}. 

Thus, real-time intraoperative delineation of cancer margins is an important step toward reducing the re-excision rate. Alternative pathology approaches such as frozen section (FS) pathology~\cite{weber1996role} and Touch Prep (TP) cytology~\cite{cox1991touch} have been proposed. However, each method exhibits limitation, preventing its wide adoption in the operating room~\cite{esbona2012intraoperative}. Therefore, surgeons have an urgent need for a method to detect the entire cancer margin intraoperatively in real time. Such a method can be divided in two parts; real-time measurement and real-time evaluation. We adopt for the first part optical coherence tomography and for the second deep neural networks.

Optical coherence tomography (OCT) is one of the promising techniques for in situ tumor margin detection of breast tissue. OCT is a high-resolution optical imaging modality capable of generating real-time cross-sectional images of tissue microstructures~\cite{fercher2003optical}. In operation, it is analogous to ultrasound imaging except that it utilizes near-infrared light waves instead of sound waves. Since its introduction, remarkable technological advances in OCT have been made, resulting in significant improvements in acquisition speeds~\cite{chen2005spectral}, spatial resolution, and added functional modalities~\cite{de2002review,makita2006optical}. OCT has also found clinical applications and commercial success in ophthalmology, cardiology, gastroenterology, and oncology~\cite{fercher2003optical}. 

OCT-based evaluation of breast cancer has been reported in many previous works. For instances, intraoperative assessment of human breast tumor margins has been reported, demonstrating high correlation of OCT images and histopathology examinations (100\% sensitivity and 82\% specificity)~\cite{bc-oct1}. Development of image processing algorithms has also been pursued to enable automated differentiation of tumor cells. Zysk et al. presented a Fourier-domain approach that detects periodicities arising from unique structural features of normal breast tissues~\cite{bc-oct2}. Mujat et al. and Savastru et al. devised methods that can differentiate tumor legions based on multiple parameters extracted from OCT A-line information. Based on the OCT images acquired with animals, the method could detect cancer-positive margins with at least 81\% sensitivity and 89\% specificity~\cite{bc-oct3}. With human tissue, it yielded 88\% sensitivity and specificity~\cite{mujat2009automated}.

As machine learning methods tend to outperform classical statistics in many image analysis applications, we investigate in this paper if such methods can improve the accuracy of real-time OCT image analysis. We report on a novel tumor margin detection method based on machine-learning analysis of OCT images of breast tissue. Our strategy is to use Deep Neural Networks (DNNs) trained using weight decay alone, weight decay and DropOut, and newly introduced function norm regularizers for analysis. The best performance is obtained by one of the new regularizers, which yields a 92\% sensitivity and a 96\% specificity as reported in Table~\ref{TAB_mean}.

In the following, we first introduce the imaging protocol, the data analysis proposed methods, and the used experimentation strategy in Sec.~\ref{Sec:method}. Then, we present the settings and results of our experiments in Sec.~\ref{Sec:experiment}. Finally, we discuss the obtained results in Sec.~\ref{Sec:discussion} and conclude in Sec.~\ref{Sec:conclusion}.

\section{Materials and method} \label{Sec:method}
The first two subsections, \ref{sec:OCTinstrument} and \ref{sec:ImagingProtocol}, explain data acquisition for our experiments.  From Subsection \ref{sec:DataAnalysisMethod} to \ref{sec:ModelSelection} we describe our novel margin assessment method using deep neural networks.  Subsection \ref{sec:DataPreparation} describes our data preparation.  Subsection \ref{sec:proc} explains the evaluation method in this paper.
 
\subsection{OCT instrument}\label{sec:OCTinstrument}
\begin{figure*}
\centering
\includegraphics[width =0.75\textwidth]{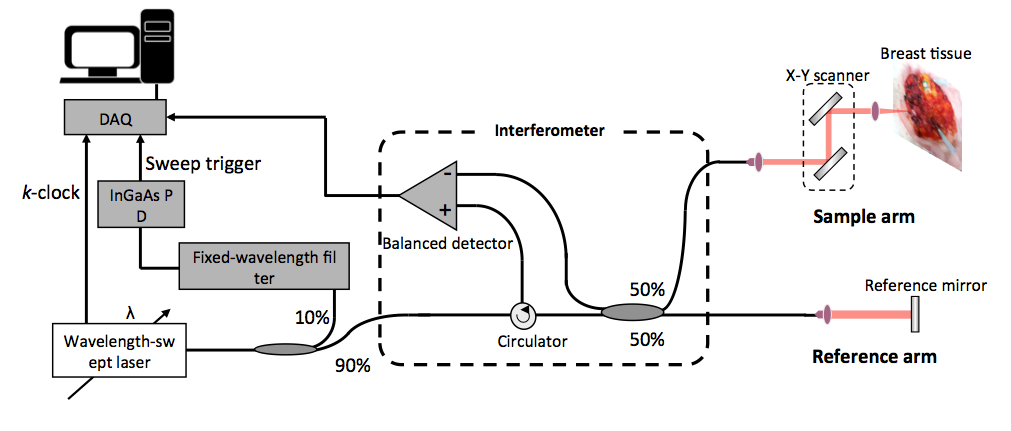}
\caption{Imaging breast tissue with OCT system.  Details of the instrument design are give in Subsection~\ref{sec:OCTinstrument}.}
\label{sys}
\end{figure*}

A custom-built OCT system was employed to image human breast tissue specimens, illustrated in Figure~\ref{sys}. The specification is as follows: The OCT system was built on optical frequency domain imaging~\cite{yun2003high,yun2003high2}, and employs a 1.31-$\mu$m wavelength-swept laser (SS-1310, Axsun Technologies, Inc., Billerica, MA, USA) as a light source. Light from the laser was first directed to a 90/10 fiber coupler. Then, 10\% of the light was guided and detected with a fast InGaAs photodetector (InGaAs-PD) through a narrowband fixed-wavelength filter. The detector generated a pulse when the laser output swept through the passband of the filter. This pulse was converted to a transistor-transistor logic pulse and fed to a high-speed digitizer (DAQ; ATS9350, AlazarTech, Pointe-Claire, QC, Canada) to trigger signal acquisition. The k-clock from the laser was directly connected to the digitizer to serve as a sampling clock. 90\% of the remaining light was directed to a Michelson-type interferometer that consisted of $2\times2$ fiber coupler, circulators, and balanced detector. The light in the sample arm passed through a two-dimensional galvanometric beam scanner. A 50-mm focal length achromatic lens in the sample arm focused 1.5 mW light onto the specimen with a beam diameter of 35 $\mu$m. The axial resolution was measured to be 5.2 $\mu$m in tissue. Reflected light from the sample and reference mirror was coupled into the interferometer, and subsequently measured by a balanced detector. The employed OCT imaging system acquires images at a rate of 50,000 axial scans/s. OCT beam scans laterally over the sample using two galvanometric beam scanners. 
The pixel resolution is 3.66 times higher in the horizontal direction than the vertical direction. Acquisition, computation and display of OCT images were performed via a custom software implemented in Visual C++.

Finally, the acquisition process takes about 0.018 sec per OCT B-scan image, and the processing of images to make them exploitable for analysis takes 2 sec per image, which makes an intraoperative application possible. A speedup is also possible using parallel GPU computing. 
\subsection{Imaging protocol}\label{sec:ImagingProtocol}

The breast tissue specimens were obtained from the Pathology Department, Severance Hospital (Seoul, South Korea). No information about tissue donors was provided. Tissue procurement, handling, and data collection were performed according to an approved Institutional Review Board protocol. Our tissue measurement protocol consisted of OCT B-scans over small areas of each sample (3-mm $\times$ 3-mm lateral scans). Each tissue sample was kept hydrated in saline solution at 37$^{\circ}$C during the measurements. After completion of OCT measurements, each tissue sample was marked with India ink on the OCT imaging locations (usually 3 to 5 locations on each sample) and fixed with formalin (10\% formalin in PBS). Histologic preparation of each tissue specimen was then performed at the histology department.

\begin{figure*}[ht]
\centering
\subfigure[OCT image of adipose tissue]{
\includegraphics[width=0.7\textwidth]{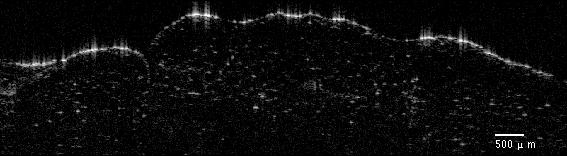} }
\hfill
\subfigure[OCT image of tumorous tissue]{
\includegraphics[width=0.7\textwidth]{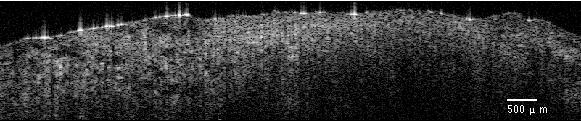} }
\caption{Breast OCT images.}
\label{fig_n_c}
\end{figure*}

Typical OCT images of normal and tumor tissue types are shown in Figure~\ref{fig_n_c}. The OCT image was cropped to 1.5 mm (horizontal) $\times$ 0.75 mm (axial) to show the tissue part of the image. A clear structural difference can be observed between the adipose tissue and the tumorous tissue: As the adipose tissue is composed of relatively large cells, a kind of periodicity is observed in the related images. The images of timorous tissue have less clear spatial structure.

\subsection{Data analysis method}\label{sec:DataAnalysisMethod}

\subsubsection{DNNs}

To analyze the data, we apply DNNs as a machine learning tool. 
Neural networks in general can be seen as directed graphs where each node is characterized by an activation function and each of the edges is characterized by a weight (see Figure~\ref{ffnet}). The activation functions are usually non-linear functions that are close to 0 (the neuron is non-activated) for some values of the input, and have non-zero values (the neuron is activated) for other inputs. Through its activations and weights $W$, the network defines a function $f_W$ that relates the inputs (images) to the outputs (unnormalized class probabilities) \cite{Goodfellow-et-al-2016-Book}.

\def\layersep{2cm}
\begin{figure}
\centering
\begin{tikzpicture}[shorten >=1pt,->,draw=black!50, node distance=\layersep]
    \tikzstyle{every pin edge}=[<-,shorten <=1pt]
    \tikzstyle{neuron}=[circle,fill=black!25,minimum size=17pt,inner sep=0pt]
    \tikzstyle{input neuron}=[neuron, fill=green!50];
    \tikzstyle{output neuron}=[neuron, fill=red!50];
    \tikzstyle{hidden neuron}=[neuron, fill=blue!50];
    \tikzstyle{annot} = [text width=4em, text centered]

    \foreach \name / \y in {1,...,4}
        \node[input neuron, pin=left:Input \#\y] (I-\name) at (0,-\y) {};

    \foreach \name / \y in {1,...,5}
        \path[yshift=0.5cm]
            node[hidden neuron] (H-\name) at (\layersep,-\y cm) {};

    \foreach \name / \y in {1,...,3}
    \node[output neuron,pin={[pin edge={->}]right:Output \#\y}, right of=H-3] (O-\name) at (\layersep,-\y) {}; {};

    \foreach \source in {1,...,4}
        \foreach \dest in {1,...,5}
            \path (I-\source) edge (H-\dest);

    
    \foreach \source in {1,...,5}
   		  \foreach \dest in {1,...,3}
      		  \path (H-\source) edge (O-\dest);

    \node[annot,above of=H-1, node distance=1cm] (hl) {Hidden layer};
    \node[annot,left of=hl] {Input layer};
    \node[annot,right of=hl] {Output layer};
    
\end{tikzpicture}
\caption{Schematic of a feed-forward neural network.}\label{ffnet}
\end{figure}
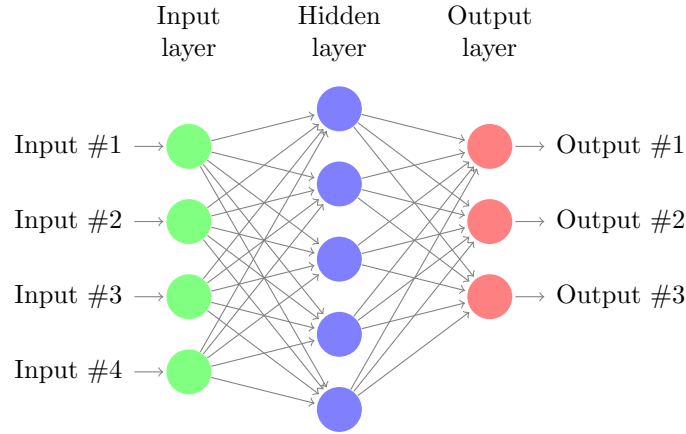

\subsubsection{Choice of DNN architecture}
In image processing, the state-of-the-art architectures are deep Convolutional Neural Networks (ConvNets) \cite{lecun1989backpropagation, lecun1995convolutional,krizhevsky2012imagenet,simonyan2014very}. These architectures have yielded the best performance on top benchmark evaluations in a large number of tasks related to image processing, including image classification~\cite{ILSVRC15}, object recognition~\cite{NIPS2013_5207}, and segmentation~\cite{crfasrnn_iccv2015}. ConvNets are neural networks that
give more importance to spatial information and proximity 
than to the global connectivity we can obtain with a fully connected network (see Figure~\ref{ffnet}). A ConvNet architecture is usually composed of the following blocks:
\begin{itemize}
\item \textbf{Convolution layer:} composed of a bank of learned filters, to be convoluted with the input images. They operate a projection of the inputs to an over-complete basis.
\item \textbf{Non-linearity:} used to sparsify the over-complete features; frequently the rectified linear unit (ReLU: $f(x) = \max(0,x)$) is used in ConvNets.
\item \textbf{Pooling:} downsampling; usually either max pooling or average pooling are used in ConvNets.
\item \textbf{Fully-connected layers:} applied after some blocks of convolution-ReLU-pooling layers. While convolution blocks operate like feature extractors, the fully connected layers operate as a classifier.
\end{itemize}

As the designed system is to be used in real time during surgery, the computation time is very important. In order to have a  test time fast enough for clinical practice, we avoid very deep networks. The LeNet-5 architecture~\cite{lecun1995learning} has a very good accuracy on small images such as MNIST~\cite{lecun1998mnist} (handwritten digits) and CIFAR~\cite{krizhevsky2009learning} (small natural images) databases. For a better localization of the tumor, we propose to divide the images in small patches, and the LeNet-5 architecture is then a reasonable choice. 

The network is composed of three convolution-ReLU-pooling layers, two fully connected layers and uses the cross-entropy loss function. To use this loss, we first estimate class probabilities by $\hat{P_i} = \frac{e^{\hat{y_i}}}{\sum e^{\hat{y_i}}}$ where $\hat{y_i}$ is the estimated output of the network, then compute the logistic loss by $-\langle T,\log(\hat{P})\rangle$ where $T$ is the indicator vector for the true class with 1 only in the true label position and 0 elsewhere, and $\log$ is applied entry-wise. 

\begin{figure*}
\includegraphics[width=0.9\textwidth]{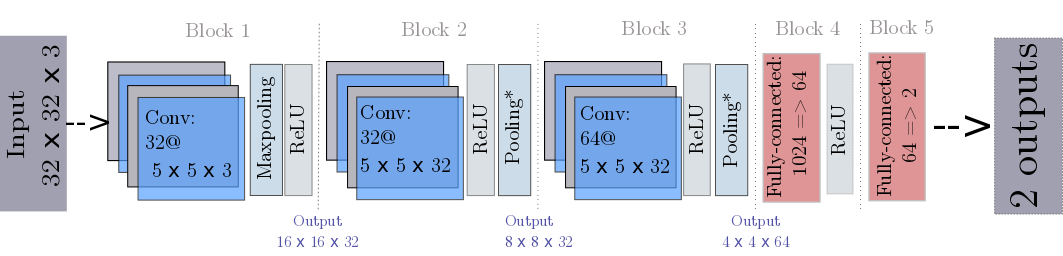}
\caption{The CNN architecture used for OCT margin assessment.}
\label{figArch}
\end{figure*}

Figure~\ref{figArch} shows a diagram of the described architecture. In this figure, the blocks noted Pooling* will be replaced by either average or max pooling blocks. The final choice for each method will be chosen by model selection. The two outputs of the network are the probabilities of the patch being tumorous (first output) or healthy tissue (second output).

\subsection{Training process}
In general, training a DNN is minimizing the empirical risk:
\begin{equation}
\mathcal{R}_N(W) = \frac{1}{N}\sum_{i=1}^N \ell(f_W(x_i),y_i)
\end{equation}
where $\ell$ is the loss function. However, DNNs depend on a large number of parameters (the weights), and require a large database for accurate minimization of $\mathcal{R}_N$. When only a small database is available, regularization is needed to avoid overfitting. The regularized problem can be written as follows:
\begin{equation}
\arg\min_{W} \mathcal{R}_N(W) + \lambda\Omega(f_W),
\end{equation}
where $\Omega$ is a measure of the function complexity.

In our case, as the possibility to collect a large database is fairly limited, regularization is required. To avoid overfitting, we consider the following four regularizers:
\begin{itemize}
\item Weight decay (WD) \cite{Goodfellow-et-al-2016-Book} : $\Omega(f_W) = \|W\|_2^2$;
\item Weight decay + DropOut (WD+DO) \cite{hinton2012improving} : $\Omega(f_W) = \|W\|_2^2$ + an implicit term introduced by model averaging \cite[Section~4]{wager2013adaptive}: at each step, a fixed ratio of randomly chosen activations  is set to 0.
\item Function norm regularization using the data distribution (FN-DD) \cite{DBLP:journals/corr/TrikiB16} : $\Omega(f_W) = \frac{1}{M} \sum_{i=1}^M \langle f_W(x_i),f_W(x_i)\rangle \rightarrow \int \langle f(x), f(x) \rangle dP(x)$, where $x_i$ are (unlabeled) images that are not used in the empirical risk.
\item Function norm regularization using slice sampling (FN-SS) \cite{DBLP:journals/corr/TrikiB16}: $\Omega(f_W) = \frac{1}{M} \sum_{i=1}^M \langle f_W(x_i),f_W(x_i)\rangle$, where $x_i$ are sampled from a distribution proportional to $\|f_W(x)\|_2^2$ using slice sampling~\cite{neal2003}.
\end{itemize}
Optimization for the last two methods is performed using \cite[Algorithm~1]{DBLP:journals/corr/TrikiB16}.  

\subsection{Model selection}\label{sec:ModelSelection}
Each of the tested methods listed above depends on at least one parameter. 
All of them use a positive scalar to control the amount of regularization. We call this parameter $\lambda_{WD} \in \mathbb{R}_+$ for WD and WD+DO and $\lambda_{FN} \in \mathbb{R}_+$ for the function norm regularization methods. In addition, WD+DO relies on a rate that designates the ratio of neurons to be dropped at each training step. This rate is denoted by $R_{DO} \in (0,1)$. Each choice for these parameters yields a different model. For each method, we test 16 models: 
\begin{itemize}
\item For WD, FN-DD and FN-SS, we test eight values for $\lambda_{WD}$ and $\lambda_{FN}$:  $10^{-5}$, $10^{-4}$, $10^{-3}$, $10^{-2}$, $10^{-1}$, 1, 10, $10^{2}$.
\item For WD+DO, we use $\lambda_{WD} = 10^{-4}$ or $10^{-2}$. For each of the two values, we test four values for $R_{DO}$: 0.1, 0.25, 0.5 and 0.75. 
\item Finally, for all methods and all models, we test both Max pooling and Average pooling for the designated blocks in Figure~\ref{figArch}.
\end{itemize}
Model selection is then operated by 5-fold cross-validation, selecting the model with the best mean area under the Receiver Operating Characteristic (ROC) curve (AUC). 

\subsection{Data preparation}\label{sec:DataPreparation}
Ultimately,  the aim of the proposed algorithm is to separate the regions that are healthy and those that are tumorous in an image of mixed tissues.  In order to do that, our strategy is to divide the images into small patches of tissue, and to train our network in order to recognize each of the types.

\begin{figure}[ht]
\centering
\subfigure[Tumorous tissue]{
\includegraphics[width= 0.7\textwidth]{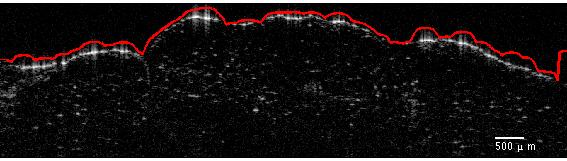}}
\subfigure[Adipose tissue]{
\includegraphics[width= 0.7\textwidth]{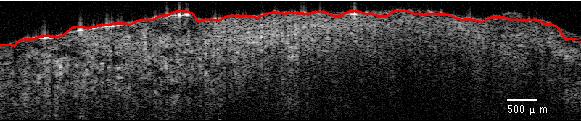}}
\caption{Surface detection using the method described in Section~\ref{sec:DataPreparation}.}
\label{surf}
\end{figure}

To extract the patches that compose our dataset, we begin with images containing only one type of tissue. These images are acquired from tissues that are separated by pathologists into tumorous and normal tissues. From these images, we need to exclude the regions corresponding to air and those that are too distant from the tissue surface, as these regions are not useful for margin assessment. This requires an accurate detection of the surface of the tissue, which is achieved using a method built on the Sobel edge detector~\cite{kanopoulos1988design}:  
\begin{enumerate}
\item A Gaussian filter (size $10 \times 10$, standard deviation 3) is applied.
\item The Sobel edge detector is applied.
\item The first pixel detected as part of the edge in each vertical line is selected. For the lines where no pixel is detected as part of the edge, the same axial position as in the previous line is taken. If no pixel in the first line is detected as part of the edge, we select the position corresponding to half of the image depth.
\item The columns are divided into ten sets by assigning the column of index $j$ to the set $j \bmod 10$. For each set, we apply a cubic spline to interpolate the tissue surface estimate, and we then average the ten estimated curves. This operation is useful to limit the effect of outliers.  
\item The obtained curve is smoothed by a ``rolling ball'' under the surface \cite{sternberg1983biomedical}. 
\end{enumerate}
Figure~\ref{surf} shows an example of the described detection. 
To ensure that no remaining pixels corresponding to air are included, the detected boundary is shifted down by 30 pixels corresponding to approximately 0.1 mm (cf.\ Section~\ref{sec:OCTinstrument}).

\subsection{Evaluation metrics}
\label{sec:proc}
Our database is extracted from three samples taken from two different patients separated beforehand into tumorous and healthy tissue by pathologists, and one sample taken from a third patient that is unlabeled (to be used for regularization in FN-DD, as well as for qualitative tests). We denote the two samples taken from the first patient by A and B, and the sample taken from the second patient by C. For each sample, a number of adjacent frames (as shown in Figure~\ref{sys}) is acquired (Table~\ref{fr_nb}). 

\begin{table}
\centering
\caption{Number of acquired frames}
\label{fr_nb}
\begin{tabular}{lll}
\toprule
& Tumor & Normal \\
\midrule
Sample A & 849 & 591 \\
Sample B & 888 & 591 \\
Sample C & 549 & 729 \\
Regularization & \multicolumn{2}{c}{664} \\
Qualitative test &\multicolumn{2}{c}{60}\\
\bottomrule
\end{tabular}
\end{table}

These images are afterwards divided into volumes of size $64\times 64\times 3$, that are downscaled to $32\times 32\times 3$ before applying the network.\footnote{The patches extracted for training do not overlap, while those extracted for test do.} This allows the localization of the tumor in three dimensions. The size of the patches is chosen to be large enough to capture the structure of the tissue, but small enough to allow an accurate localization of the tumor.  To have a reliable estimation of the performance of each of the methods, the following protocol is employed:
\begin{enumerate}
\item We consider four database configurations: \begin{inparaenum}[(i)] \item train on A and test on C, \item train on B and test on C, \item train on C and test on A, and \item train on C and test on B\end{inparaenum}. Using these data splits, cross-patient generalization is evaluated.  
Table~\ref{nb_patches} gives the number of training, test and regularization patches for each configuration.
\item For each of the configurations and each of the methods, the best model is chosen by 5-fold cross-validation.
\item The selected model is used to estimate the class probabilities of each test patch.
\item The highest output of the network yields the estimated class for each of the patches in the test set. We then count the number of true positives ($TP$), true negatives ($TN$), false positives ($FP$) and false negatives ($FN$) by comparing our classification to the ground truth provided by the pathologists.  
\item The results obtained for all the configurations are then averaged to provide an estimate of the expected performance of the studied methods.
\item These results are finally compared to the performance reported in state-of-the-art works.
\end{enumerate}
We also show qualitative results in Figure~\ref{qual1}. 
\begin{table}
\centering
\caption{Number of patches per database configuration}
\label{nb_patches}
\begin{tabular}{ll|l|l|l|l}
\toprule
\multicolumn{2}{l|}{Patches} & Train : a & Train : b & Train : c & Train : c\\
\multicolumn{2}{l|}{} & Test : c & Test : c & Test : a & Test : b\\
\midrule
Train & Cancer&2,830 &3,552 &  \multicolumn{2}{c}{1,158}\\
& Normal & 2,364& 1,970& \multicolumn{2}{c}{2,411}\\
\midrule
Test & Cancer&\multicolumn{2}{c|}{10,615}& 26,885&35,520 \\
& Normal &\multicolumn{2}{c|}{24,117} & 21,670&18,715\\
\midrule
\multicolumn{2}{l|}{Regularization} & \multicolumn{4}{c}{6,413}\\
\bottomrule
\end{tabular}
\end{table}

Using $TP$, $TN$, $FP$ and $FN$ computed on the test data in each of the four data  configurations, five measurements are computed, Sensitivity ($Se$), Specificity ($Sp$), Precision ($Pr$),\footnote{Precision and recall are typically reported together.  We note that recall is identical to the notion of sensitivity.} the F1-score ($F1$), the G-mean ($G$), and the Matthews Correlation Coefficient ($MCC$):
\begin{align}
&Se = \frac{TP}{TP+FN},\ Sp = \frac{TN}{TN+FP}, \ Pr=\frac{TP}{TP+FP} \nonumber\\
&F1 = \frac{2\times Pr\times Se}{Pr + Se},\ G = \sqrt{Se\times Sp} \nonumber\\
&MCC = \frac{\frac{TP}{N} - S\times P}{\sqrt{P\times S\times (1-P)\times (1-S)}}
\end{align}
where $N=TP+TN+FP+FN$ is the total number of pixels of the image, $S=(TP+FN)/N$, and $P=(TP+FP)/N$. 

Sensitivity ($Se$) measures the capability of the method to correctly detect tumor patches, while specificity ($Sp$) indicates its capability to correctly detect the healthy patches. Though a higher $Se$ value is desired, it is important to analyze it jointly with $Sp$ as it is possible to have $Se = 1$ by considering trivially all the patches as tumorous. Moreover, a high $Sp$ means a reasonable preservation of the normal tissue, which is important for the patient outcome. Precision ($Pr$) quantifies the ratio of patches classified as cancer that are correctly identified. As our data may be imbalanced (the normal tissue tends to have a higher volume than the cancer mass), we include the F1-score ($F1$), the G-mean ($G$)~\cite{he2009learning}, and the
Matthews Correlation Coefficient ($MCC$)~\cite{matthews1975comparison}. 
The $F1$-score is the harmonic mean of precision and sensitivity. It achieves its maximum value of 1 when the detection of the positive class is perfect, and its lowest value of 0 when the classification is completely wrong. Similarly, the G-mean gives the balance between $Se$ and $Sp$ by taking their geometric mean, returning a value between 0 and 1. The $MCC$
is a correlation coefficient between the ground truth and the predicted classes. It returns a value between -1 and +1, with
+1 indicating a perfect prediction, 0 no better than random, and -1 a total disagreement between prediction and ground truth. 

Finally, receiver operating characteristic (ROC) curves are also generated from the normalized probability estimates from the softmax layer of the network at test time, 
and the area under each curve ($AUC$) is computed. In addition to these measures, we report the overall accuracy at the selected threshold ($ACC$), the equal error rate ($EER$) which corresponds to $100 \times (1-Se) =100 \times (1-Sp) = 100-ACC$ at the point of the ROC where $Se = Sp$.

\section{Experiments and results} \label{Sec:experiment}

All the experiments were conducted using MatConvNet~\cite{vedaldi15matconvnet}, on a machine equipped with a 4 core CPU (4.00 GHz) and a GPU GeForce GTX 750 Ti.  For all settings, we train the networks during 45 epochs. The learning rates are based on an exponentially decreasing schedule following standard practice~\cite{krizhevsky2012imagenet}: 
0.05 for the 30 first epochs, 0.005 from epochs 30 to 40, and 0.0005 for the last five epochs. The momentum rate for Weight Decay and DropOut is set to 0.95 for WD and WD+DO. It is set to 0 or the FN methods because it was observed that it causes the divergence of the training process otherwise. For FN-SS, the number of generated samples for regularization is exactly the same as the number of regularization samples (see Table~\ref{nb_patches}).

As an additional baseline, we trained support vector machines (SVMs)~\cite{cortes1995support} with several types of kernels~\cite{boser1992training}: linear, polynomial with degrees 2 and 3, and Gaussian with a variance equal to the median of the squared distance between the data points. On these data, the linear kernel gave the best validation set accuracy, and is reported in the sequel.

In the following subsections we report: 
\begin{enumerate}
\item The selected models for the DNNs and the evolution of the corresponding training and test top-1 error across training epochs, which represents in our case the accuracy as defined in Section~\ref{sec:proc}. 
\item Quantitative results: mean ROC curves and the averages of the measures described in Section~\ref{sec:proc} on each of the four train-test settings.  
For computation of the mean ROC curves, we used vertical averaging~\cite{fawcett2006introduction}. 
\item Qualitative results: we visualize classification results on a sample for which both normal and tumorous tissue is present. 
This visualization gives a qualitative view that can be provided to surgeons in an intraoperative setting.  We also provide an analysis of the speed of computation at test time.
\end{enumerate}

\subsection{Selected models}
For all trials and all methods, max pooling gives better results than average pooling. The parameters determined by the model selection procedure are reported in Table~\ref{selected}. In this table, we observe that for both of the function-norm methods, there is a change in the selected model when we retrain on the data from sample C. 

\begin{table}
\centering
\caption{Selected models}
\label{selected}
\begin{tabular}{llllll}
\toprule
\multicolumn{2}{c}{Method} & Train : a & Train : b & Train : c (1) &  Train : c (2)\\
\multicolumn{2}{c}{\& Parameter}&&&&\\
\midrule
WD & $\lambda_{WD}$ & $10^{-4}$ & $10^{-3}$ & $10^{-2}$ & $ 10^{-2}$ \\
WD+DO & $\lambda_{WD}$&$ 10^{-4}$ & $10^{-4}$ & $ 10^{-2}$  &$ 10^{-2}$\\
& $R_{DO}$&  0.1 & 0.1 & 0.25 & 0.25\\
FN-DD & $\lambda_{FN} $& $1$ & $10$ & $10^{-1}$ & $10^{-3}$\\
FN-SS & $\lambda_{FN} $& 10 & 10 &  1 & $10^{-3}$\\
\bottomrule
\end{tabular}
\end{table}

\subsection{Quantitative assessment}
For each of the trials, the selected model is applied to classify the designated test sample. 
The obtained performance measurements are then averaged over the four trials (cf.\ Sec.~\ref{sec:proc}). The obtained results with their corresponding error bars are displayed in Table~\ref{TAB_mean}. We also show the performance measures for state-of-the-art methods.\footnote{Performance measures for methods reported in other papers \cite{bc-oct1,mujat2009automated,bc-oct3,bc-oct2} are taken directly from those publications.  Neither data nor source code for these methods were publicly available at the time of writing this article.} From these numbers, we can see that on average, FN-SS is slightly better than all the other methods. However, the error bars suggest that this improvement is not statistically significant.

\begin{figure}
\subfigure[All methods]{
\includegraphics[width=0.45\columnwidth]{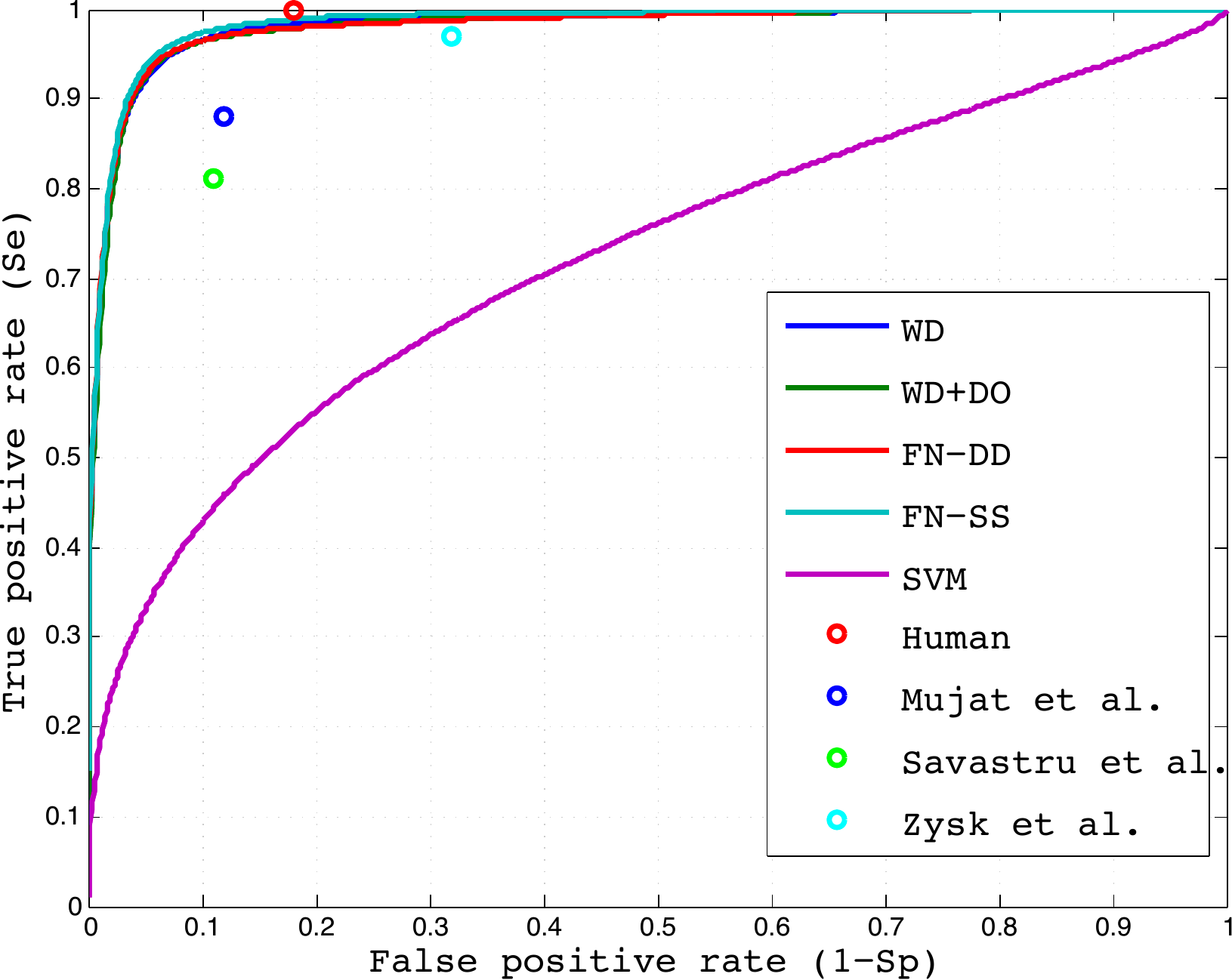}
\label{ROC1}}
\hfill
\subfigure[Zoom-in for DNN methods]{
\includegraphics[width=0.45\columnwidth]{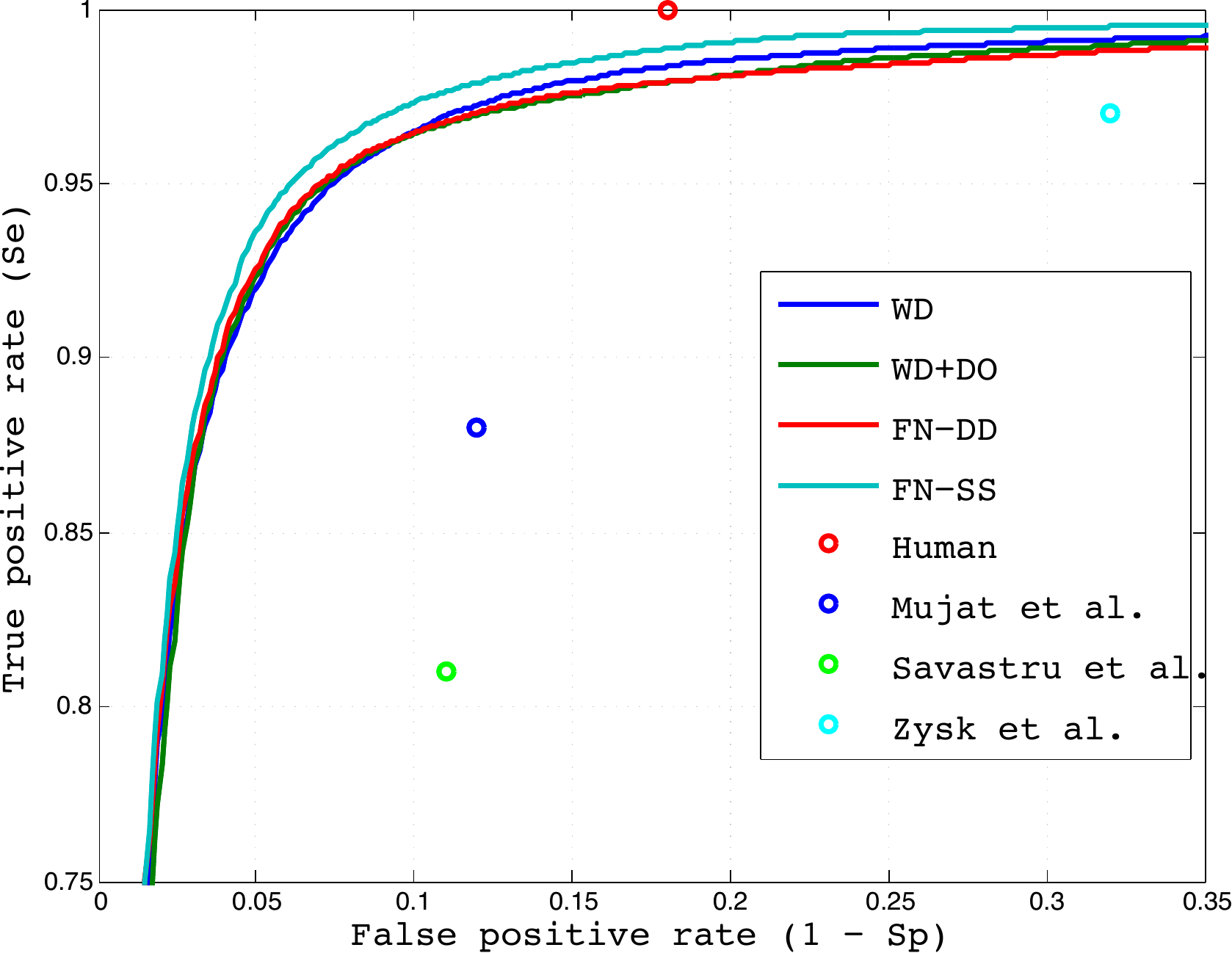}
\label{ROC2}}
\caption{Mean ROC curves for support vector machines (labeled SVM) and deep learning methods evaluated in this work.  Comparison to the existing literature is based on performances reported in the respective papers evaluated on different test data.  References for previous publications can be found in Table~\ref{TAB_mean}.}
\label{ROC}
\end{figure}

Figure~\ref{ROC} shows mean ROC curves. In Figure~\ref{ROC1}, we show the mean ROC curves over the four trials for all the tested methods. For \cite{bc-oct1,mujat2009automated,bc-oct3,bc-oct2}, we display the performance points reported in the corresponding papers, since ROCs are not accessible. For SVM, we show only the best curve in order to avoid obstructing the figure.  In Figure~\ref{ROC2}, we zoom in to better visualize the curves of the best performing methods. We can see clearly in this figure that the curves for all DNN based methods are higher than the state-of-the-art points related to automated detection, and thus closer to the human detection point (red point in the figure). 

\begin{table*}
\centering
\caption{Evaluation metrics for all the tested methods vs.\ state-of-the-art methods in the literature.  It can be seen that competing methods make different trade-offs between $Se$ and $Sp$, but that function norm regularization with slice sampling (FN-SS \cite[Sec.~2.2]{DBLP:journals/corr/TrikiB16}) dominates across standard metrics that account for a balance between true-positives and true-negatives.  This dominance is reinforced by the ROC curves in Figure~\ref{ROC}.}
\label{TAB_mean}
\resizebox{\textwidth}{!}{%
\begin{tabular}{lrlrlrlrlrlrlrlrlrl}
\toprule
Method && $Se$ && $Sp$ && $Pr$ && $F1$ && $G$ && $MCC$ && $ACC$(\%) && $AUC$ && $EER$(\%)\\
\toprule
WD &&0.8843 &&0.9667 && 0.9479 && 0.9135 &&0.9241&&  0.8462&& 92.75 && 0.9803 &&  5.83\\
&$\pm$& 0.0335&$\pm$& 0.0066&$\pm$& 0.0195&$\pm$& 0.0187&$\pm$& 0.0180&$\pm$& 0.0375&$\pm$&  2.01&$\pm$& 0.0040 &$\pm$&0.83  \\
\midrule
WD+DO &&  0.8960 && 0.9648&& \textbf{0.9480}  && 0.9204 &&0.9292&&  0.8613&& 93.72&& 0.9792 && 5.67 \\
&$\pm$& 0.0358&$\pm$ &0.0052&$\pm$ & 0.0160&$\pm$&  0.0232&$\pm$&  0.0192&$\pm$& 0.0349&$\pm$ &1.62 &$\pm$ &0.0058  &$\pm$& 1.15\\
\midrule
FN-DD && 0.9031 && \textbf{0.9651} && 0.9462 && 0.9233  &&0.9320&& 0.8701&& 94.24&& 0.9798 &&  5.66 \\
&$\pm$ & 0.0387&$\pm$& 0.0060&$\pm$& 0.0187&$\pm$& 0.0272&$\pm$& 0.0212&$\pm$& 0.0349&$\pm$&1.47 &$\pm$& 0.0078 &$\pm$& 1.54  \\
\midrule
FN-SS && 0.9171  && 0.9631 && 0.9479 && \textbf{0.9314} &&\textbf{0.9392}&&\textbf{0.8849}&& \textbf{94.96}&& \textbf{0.9837} &&   \textbf{5.22}\\
&$\pm$& 0.0363&$\pm$& 0.0083&$\pm$&0.0167&$\pm$& 0.0248&$\pm$&  0.0187 &$\pm$& 0.0293&$\pm$&1.18 &$\pm$& 0.0049 &$\pm$& 
1.21\\
\midrule
SVM && 0.6011&&0.6667 &&0.7287 && 0.6275&&0.4789 && NaN && 72.2260&& 0.7186 && 33.58\\
&$\pm$&0.1418 &$\pm$&0.2249 &$\pm$& 0.0266 &$\pm$& 0.0676 &$\pm$& 0.1610 && &$\pm$&3.0492 &$\pm$&0.0573 &$\pm$&2.13 \\
\midrule
Human classification~\cite{bc-oct1} && 1 && 0.8182 && 0.8182 && 0.9 && 0.9045 && 0.8182&&90&&-&&-\\
\midrule
Mujat \& al.~\cite{mujat2009automated} && 0.88 && 0.8750 && 0.7333 && 0.8 &&0.8775&& 0.7178&&87.64&&-& $\approx$& 12\\
\midrule
Savastru \& al.~\cite{bc-oct3} && 0.81 && 0.89 && - && - &&0.8491&&-&&-&&-&&-\\
\midrule
Zysk \& al.~\cite{bc-oct2} && \textbf{0.97} && 0.68 && - && - &&0.8122&&-&&-&&-&&-\\
\bottomrule
\end{tabular}
}
\end{table*}

\subsection{Qualitative results}\label{sec:QualitativeResults}

\begin{figure*}
\includegraphics[width =\textwidth]{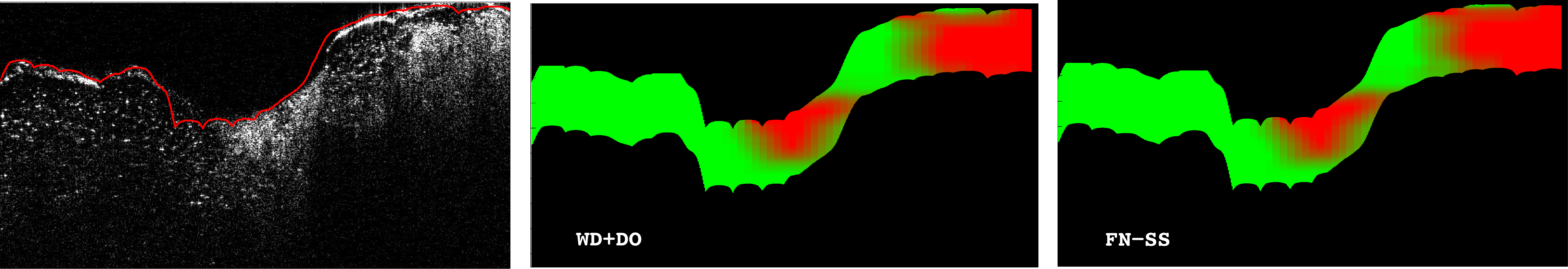}
\caption{Example OCT slice of a region with both tumorous and healthy tissue (left), as well as example detections by two deep learning variants (center and right).  Tumorous tissue is characterized by lighter regions in the OCT image.  In the center and right figures, red indicates that the deep learning system has classified the tissue as tumorous while green indicates the system has indicated the tissue is healthy. Surface detection is sufficiently accurate, and deep learning models have correctly identified the tumorous regions and their spatial extent.}
\label{qual1}
\end{figure*}

In this section, we report qualitative results. For this purpose, we used 60 slices of a 
sample that contains both healthy and tumorous tissue (Figure~\ref{qual1}). 
For these slices, we extract overlapping patches (overlap of 56 / 64 pixels in both horizontal and vertical directions, overlap of 2/3 pixels in the third dimension) after surface detection. Then, each of the patches is classified using 
the trained models. Here, we assign each patch to the class that yields the highest probability under a trained model. For each pixel, we average the predicted classes over all patches in which the pixel is included. In this manner, we obtain an average prediction for each of the pixels between $0$ (indicating healthy tissue) and $1$ (indicating tumorous tissue). For our visualization, we consider pixels that are at a depth smaller than 384 pixels below the detected surface. For margin assessment, this is the relevant depth ($> 1$ mm)~\cite{bc-oct1}. Finally, the average prediction is used to define the color of the pixel. We define:
\begin{align}
&\operatorname{green} = 1-\operatorname{prediction};\\ 
&\operatorname{red} = \operatorname{prediction} 
\end{align}
Each pixel is then colored using the RGB triplet ($\operatorname{red},\ \operatorname{green},\ 0$). The pixels outside of the useful region are colored in black. In Figure~\ref{qual1}, we show some of the obtained classifications trained on sample A. 
In these examples, we see that we can obtain very precise localization of the cancer tissue using deep learning methods.

We note that in an intraoperative setting, it may not be necessary to exactly delineate the boundaries of a tumorous region as long as we detect some true positives per region. 
This alert can be sufficient in practice as it will indicate a positive margin and attract the attention of the surgeon. 
We therefore develop a quantitative evaluation for this setting. We applied the same test procedure described above to each of the 60 images, and we assigned to each of the obtained results a score for the quality of detection of the cancer parts and the normal parts. For each of the cancer parts, we first divide the pixels into 4 subsets: pixels for which  $\operatorname{red} \geq 0.75$, pixels for which  $0.5 \leq \operatorname{red} \leq 0.75$, pixels for which  $0.25 \leq \operatorname{red} \leq 0.5$, and pixels for which  $\operatorname{red} \leq 0.25$. Then, we assign to each of the cancer regions:
\begin{itemize}
\item 1 if the first set has at least 300 pixels
\item 0.75 if the union of the first and second sets has at least 300 pixels
\item 0.5 if the union of the first, second and third sets has at least 300 pixels
\item 0 otherwise
\end{itemize}
Note that this assessment will consider the presence of a perceivable alert in the region, and not the quality of segmentation. The two first sets will indicate a perceivable region with bare eyes, while the third and last sets will be greenish regions in the colored images, but can be indicated with a simple computation.  For each of the images, the mean score over the cancer regions 
is computed. The ground truth of the segmented cancer region is produced approximatively from the results shown in Figure~\ref{qual1}. In this manner, each image will have a score between 0 and 1 for the cancer detection. The same measure computed on the healthy region gives an indication of the quality of conservation of the normal tissue: the lowest measure is the best in this case. Note that a value higher than 0.5 indicates that the considered region is more likely to be tumorous, while a measure lower than 0.5 indicated that the considered region is more likely to be healthy.  

As this score indicates some kind of sensitivity but relies on the subjective analysis of the surgeon, we call it \textit{subjective sensitivity}.   
Table~\ref{subjSe} shows the mean subjective sensitivity of the cancer regions for models trained on each of the samples (A, B and C), and over all the estimated classifications in along with the same measures for the healthy regions.

\begin{figure}
\centering
\includegraphics[width=0.4\textwidth]{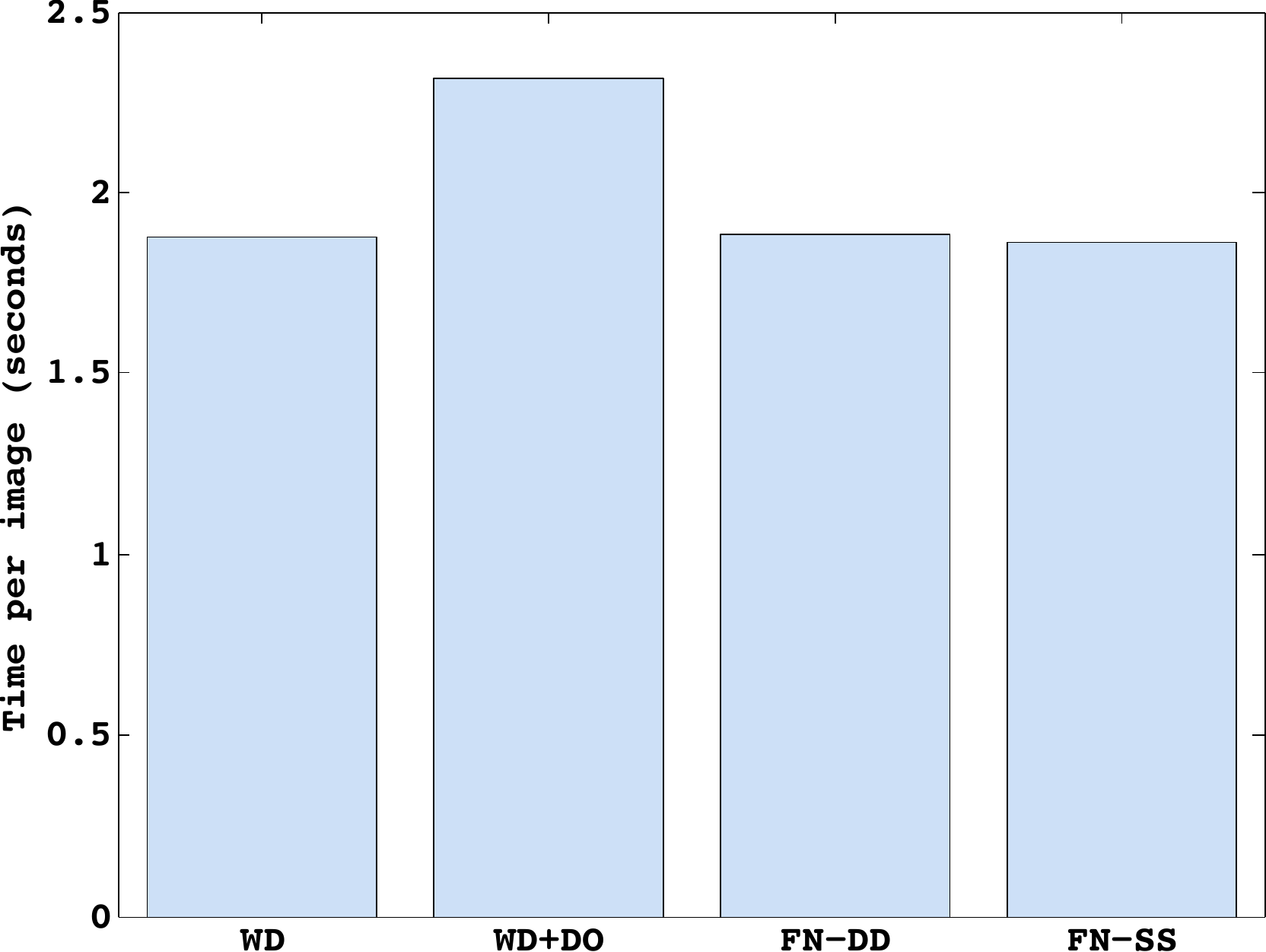}
\caption{Test time per image for the DNN models trained with various regularization strategies.  Each method takes approximately 2 seconds per image and is therefore feasible to apply in an intraoperative setting.}
\label{time}
\end{figure}

Finally, we also report here the required time to obtain results for an image in a real application situation, i.e.\ using already trained models. Figure~\ref{time} shows that all of the methods require approximately two seconds to estimate the class of each pixel of the image using the procedure described above.  As this test procedure is embarrassingly parallel, and can be performed at the same time as image acquisition, this is feasible for intraoperative application.

\begin{table}
\centering
\caption{Mean subjective sensitivity measured for cancer and normal regions in a mixed sample across three models trained on samples A, B and C and 60 slices. While the difference in performance for the measures related to cancer regions are not significant, the method FN-SS has a significantly better performance in preserving the normal tissue.}
\label{subjSe}
\begin{tabular}{lll}
\toprule
Method & Cancer regions & Normal regions \\
\midrule
WD &  0.7792 & 0.25\\
&$\pm$ 0.0188 &$\pm$ 0.0187  \\
WD+DO & 0.6618 & 0.2653\\
&$\pm$ 0.0287 &$\pm$ 0.0230 \\
FN-DD & 0.7882 &  0.2708\\
&$\pm$  0.0170 &$\pm$ 0.0249 \\
FN-SS & 0.7944 & \textbf{0.1833}\\
&$\pm$ 0.0162 & \textbf{$\pm$  0.018 }\\
\bottomrule
\end{tabular}
\end{table}

\section{Discussion} \label{Sec:discussion}

In this paper, we applied DNNs for breast cancer margin assessment with state-of-the-art regularization methods, including weight decay, weight decay + DropOut, function norm regularization based on a data distribution estimated from unlabeled samples, and function norm regularization based on a slice sampling procedure. 

The quantitative results on separated data show that the new approach for margin assessment based on DNNs reports quantitative metrics substantially higher than those previously reported in the literature (Table~\ref{TAB_mean}). However, this comparison has been made only on the base of the reported numbers in the related papers, and not by applying the methods on our data as the code used for the classification in these papers is not released.  
Our code is available for download from \url{https://github.com/AmalRT/DNN_Reg}. Another important remark is that the mean ROC curves from DNN methods are significantly closer to the human assessment than the previous automated detection methods (see Figure~\ref{ROC2}). In addition, when we compare the DNN methods, we see that on average, the function norm regularization methods have better performance (Table~\ref{TAB_mean}, Figure~\ref{ROC1} and Figure~\ref{ROC2}). 

The qualitative results on mixed data showed also the effectiveness of all of the DNNs methods. The methods alert to the presence of cancer in the related regions with equivalently high confidence score, indicating they tend to have a high probability of correctly detecting cancer tissue. When it comes to preserving the normal tissue, Table~\ref{subjSe} shows that the function norm regularization with slice sampling has a significantly better performance, supporting that even a small improvement, as observed in Figure~\ref{ROC2} and Table~\ref{TAB_mean}, can be beneficial in practice. For clinical application,  the use of this regularization method may result in a better patient outcome, but this requires further study. 

The results obtained in the qualitative results use images for which the contrast was enhanced manually, a process that would, strictly speaking, need to be repeated in an operative setting. It is desirable to perform this operation automatically, but this was not evaluated in this study. The results also depend on the surface detection. The relatively straightforward approach employed in this article works well in practice, but can likely be improved.
Finally, a critique of the FN-DD results is that they have employed a data sample from the same patient for regularization, which may not be available in practice. 
However, the FN-SS method uses no such additional data sample and achieves almost identical results.  We conclude from both quantitative and qualitative results that the FN-SS method is the most suitable for further study and eventual real world deployment.

The model selection procedure shows that the selected models for the function norm regularization methods can change when we retrain on the same data. However, the quality of the results on the test set suggests that this does not effect the behavior of the model significantly. A possible explanation to this observation is that our methods are robust to the choice of $\lambda$ and behave well as long as this parameter is in a certain range. This is an interesting area for further research to determine if this trend is repeated in other data samples. 

\begin{figure}
\centering
\includegraphics[width= 0.4\textwidth]{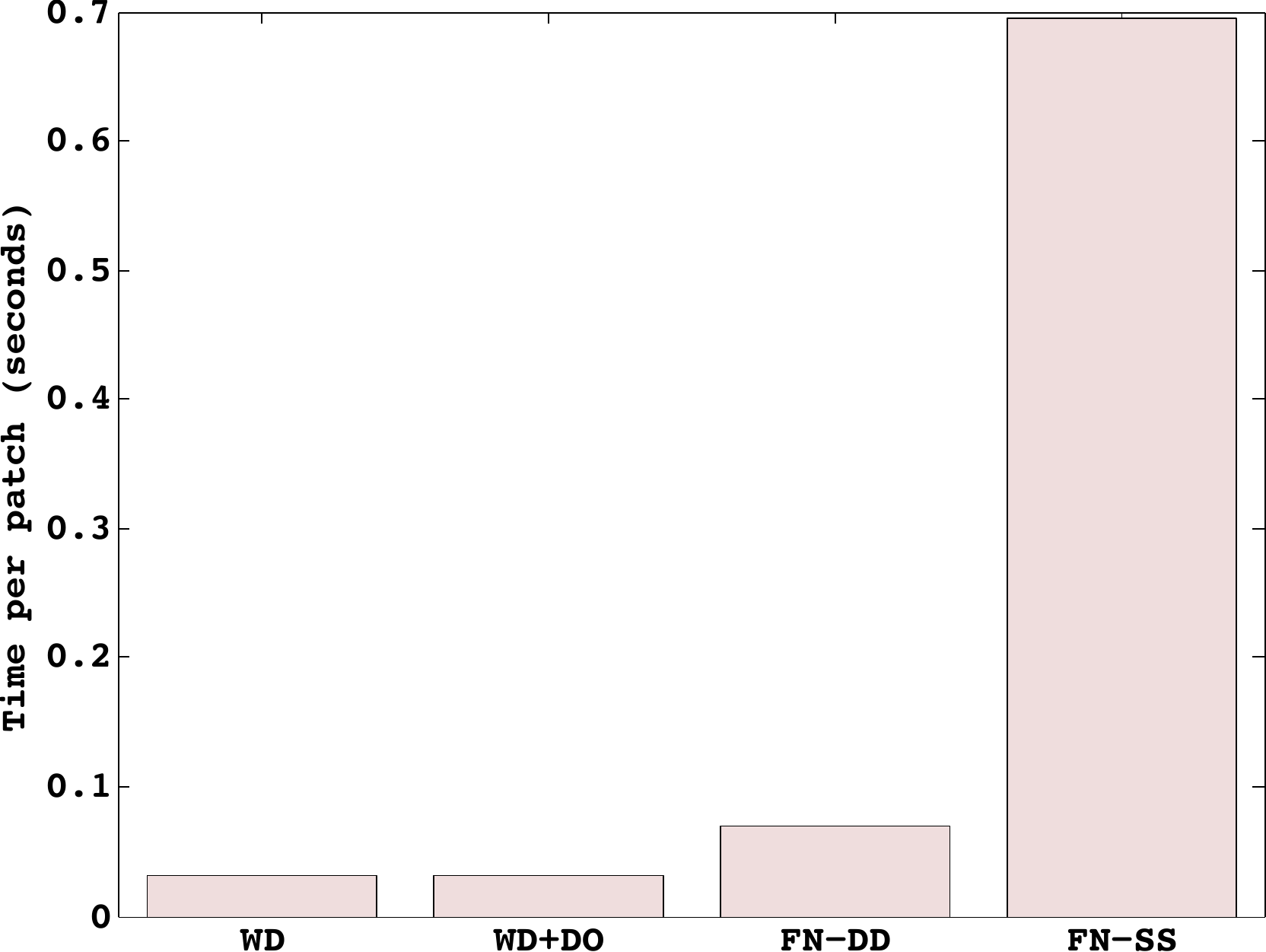}
\caption{Training time per patch. Although the total training time can be long (Training time per patch $\times$ number of patches (Table~\ref{nb_patches})), the training is done off-line. It will not affect the time needed during real time application, and the method with longer time can result in an improvement in patient outcome, while the monetary cost of additional computation is typically low~\cite{walker2009real}.}
\label{train_time}
\end{figure}

However, there is a price to pay for the high performance of the DNN methods: a longer training time. 
Most state-of-the-art methods in the literature
are based on exploiting classical statistics and are likely to be faster to train. We cannot affirm this with certainty because we do not have access to the computation time for these methods. Moreover, function norm regularization methods require a longer time to train than other DNN regularization strategies. In Figure~\ref{train_time}, we see that FN-DD requires roughly twice the time of WD and WD+DO. This can be expected as this method feeds to the network roughly twice the number of patches during training (see number of patches for training and regularization in Table~\ref{nb_patches}). FN-SS requires an even longer time because it also includes the time needed to generate the samples. Speed-up of the sampling procedure is likely possible, but this is left for future work. 
We note that the additional computational cost of these methods has a low monetary cost compared to the potential improvement in patient outcomes~\cite{walker2009real}.  The test time computation remains unaffected. 
In real time application, the required time per image is displayed in Figure~\ref{time} where we see that this time is the highest for WD+DO (between 2 and 2.5 seconds), and comparable for the three other methods (between 1.5 and 2 seconds). None of these differences is likely to have any impact on the patient in an intraoperative setting. 

\section{Conclusions} \label{Sec:conclusion}

This work implements successfully a simulation of an intraoperative margin assessment for breast tissue using OCT images. We show in this work the benefit of use of DNNs in image patch identification during surgery. We implement various regularization methods for DNN training based on function norms. We showed the efficiency and benefits of function norm regularization compared to other state-of-the-art training methods for DNNs. 

The main advantage of our training method is that it relies on a smaller number of hyperparameters than WD and WD+DO, which need tuning. Indeed, for our methods, we need only to tune the regularization parameter $\lambda$ and the batch size (fixed in this work for all methods), while for WD we need also to tune the momentum rate and for WD+DO the momentum and DropOut rates. Moreover, it is more robust to the data size and imbalance.  The impact on patient outcome is an interesting direction for future research.

\section*{Acknowledgments}

This work is partially funded by Internal Funds KU Leuven
and FP7-MC-CIG 334380 and by the research program of the National Research Foundation of Korea (NRF) (NRF-2015R1A1A1A05001548 to C. J.). Y.M. Jung has been supported by Basic Science Research Program through the National Research Foundation of Korea (NRF) of Korea (2014R1A1A2054763, 2016R1D1A1B03931337).

\newpage

\bibliographystyle{IEEEtran}
\bibliography{IEEEabrv,biblio1}

\end{document}